\PassOptionsToPackage{unicode}{hyperref}
\PassOptionsToPackage{hyphens}{url}
\documentclass[
  11pt,
]{article}
\usepackage{amsmath,amssymb}
\usepackage{iftex}
\ifPDFTeX
  \usepackage[T1]{fontenc}
  \usepackage[utf8]{inputenc}
  \usepackage{textcomp} % provide euro and other symbols
\else % if luatex or xetex
  \usepackage{unicode-math} % this also loads fontspec
  \defaultfontfeatures{Scale=MatchLowercase}
  \defaultfontfeatures[\rmfamily]{Ligatures=TeX,Scale=1}
\fi
\usepackage{lmodern}
\ifPDFTeX\else
\fi
\IfFileExists{upquote.sty}{\usepackage{upquote}}{}
\IfFileExists{microtype.sty}{% use microtype if available
  \usepackage[]{microtype}
  \UseMicrotypeSet[protrusion]{basicmath} % disable protrusion for tt fonts
}{}
\makeatletter
\@ifundefined{KOMAClassName}{% if non-KOMA class
  \IfFileExists{parskip.sty}{%
    \usepackage{parskip}
  }{% else
    \setlength{\parindent}{0pt}
    \setlength{\parskip}{6pt plus 2pt minus 1pt}}
}{% if KOMA class
  \KOMAoptions{parskip=half}}
\makeatother
\usepackage[margin=1in]{geometry}
\usepackage{authblk}
\usepackage{graphicx}
\makeatletter
\def\maxwidth{\ifdim\Gin@nat@width>\linewidth\linewidth\else\Gin@nat@width\fi}
\def\maxheight{\ifdim\Gin@nat@height>\textheight\textheight\else\Gin@nat@height\fi}
\makeatother
\setkeys{Gin}{width=\maxwidth,height=\maxheight,keepaspectratio}
\makeatletter
\def\fps@figure{htbp}
\makeatother
\providecommand{\tightlist}{%
  \setlength{\itemsep}{0pt}\setlength{\parskip}{0pt}}
\ifLuaTeX
  \usepackage{selnolig}  % disable illegal ligatures
\fi
\usepackage[numbers]{natbib}
\setcitestyle{numbers,open={[},close={]},citesep={][}}
\makeatletter
\def\NAT@def@citea{\def\@citea{\NAT@separator}}%
\def\NAT@def@citea@space{\def\@citea{\NAT@separator}}%
\def\NAT@def@citea@close{\def\@citea{\NAT@@close\NAT@separator}}%
\def\NAT@def@citea@box{\def\@citea{\NAT@mbox{\NAT@@close}\NAT@separator}}%
\makeatother
\usepackage{bookmark}
\IfFileExists{xurl.sty}{\usepackage{xurl}}{} % add URL line breaks if available
\hypersetup{
  pdftitle={The Agentification of Scientific Research: A Physicist’s Perspective},
  pdfauthor={Xiao-Liang Qi},
  hidelinks,
  pdfcreator={LaTeX via pandoc}}

\title{The Agentification of Scientific Research: A Physicist’s Perspective}
\author{Xiao-Liang Qi}
\affil{Leinweber Institute for Theoretical Physics, Stanford University, Stanford, CA 94305, USA}
\date{}

\begin{document}
\maketitle

\begin{abstract}
This article argues that the most important significance of the AI
revolution, especially the rise of large language models, lies not
simply in automation, but in a fundamental change in how complex
information and human know-how are carried, replicated, and shared. From
this perspective, AI for Science is especially important because it may
transform not only the efficiency of research, but also the structure of
scientific collaboration, discovery, publishing, and evaluation. The
article outlines a gradual path from AI as a research tool to AI as a
scientific collaborator, and discusses how AI is likely to fundamentally
reshape scientific publication. It also argues that continuous learning
and diversity of ideas are essential if AI is to play a meaningful role
in original scientific discovery.
\end{abstract}

\section{Background: The Impact of Large Language
Models}\label{background-the-impact-of-large-language-models}

Deep neural network-based AI has developed rapidly over the past decade,
but the revolution brought by Large Language Models (LLMs) is
particularly profound compared to previous advancements. In modern
physics, information is increasingly recognized as fundamental,
potentially serving as the underlying concept behind the laws of
spacetime and matter. We believe the nature of this new AI revolution can
be viewed through the lens of information.

When examining a complex system, the key lies in how its most critical
information is controlled, carried, and processed. In other words, we
should ask how the most complex information processing is accomplished,
what carrier it relies on, and how that carrier changes the dynamics of
the system as a whole. From this perspective, the present AI revolution
is not merely a technological upgrade. It is a new stage in the history
of information dynamics on Earth.

\subsection{Three Major Transformations of Information
Dynamics}\label{three-major-transformations-of-information-dynamics}

\begin{figure}
\centering
\includegraphics{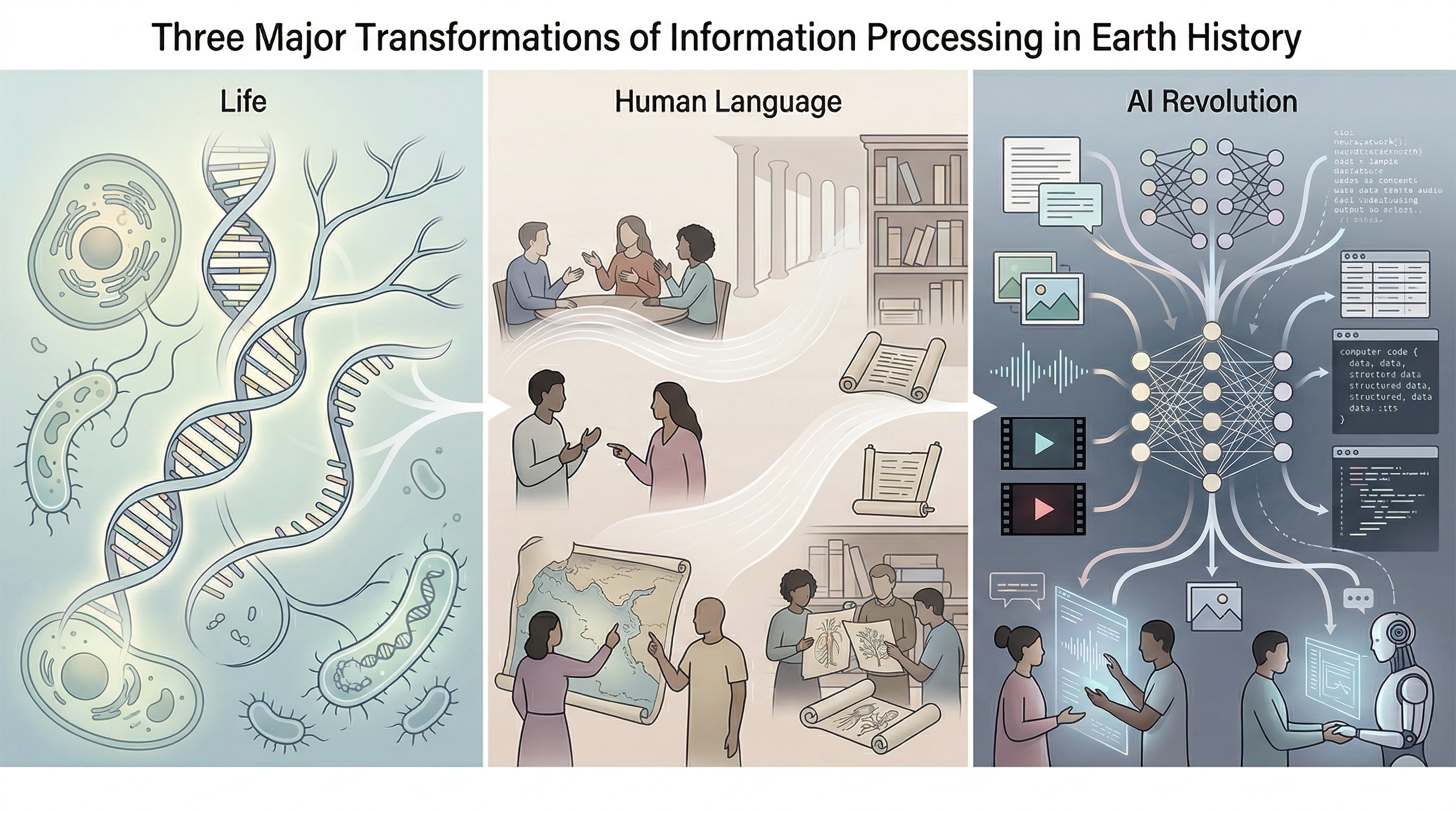}
\caption{Illustration of the three major transformations of information
dynamics in Earth's history: life, human language, and the AI
revolution.}
\end{figure}

In Earth's history, the dominant carrier and processor of complex
information has undergone several major transformations. Each
transformation changed not only the speed of information transmission,
but also the structure of adaptation, learning, and collective
evolution.

The first major transformation was the emergence of life. With DNA and
RNA as information carriers, the traits and behaviors of living systems
could be stored, copied, and modified across generations. Biological
evolution became possible because useful information was no longer lost
with the death of an individual organism. Instead, it could be
preserved, accumulated, and refined over long timescales. In this stage,
the replication and processing of complex information were embedded in
life itself.

The second major transformation was the emergence of human language.
Language made it possible for experience, memory, and knowledge to be
transmitted directly between individuals and across generations without
waiting for biological inheritance. Compared with genetic evolution,
linguistic and cultural evolution proceeded at a dramatically faster
pace. Human societies could accumulate ideas, institutions, and
technologies through communication, education, and writing. In this
sense, the most important information dynamics in human civilization
shifted from DNA-based evolution to language-based cultural evolution.

The third major transformation is the current AI revolution. Starting
from human language, AI models are increasingly able to represent and
process many different forms of information, including text, images,
audio, video, and structured data, within a unified framework. This
means that for the first time in history, the most complex forms of
information processing are no longer the exclusive domain of the human
brain. Previous information technology revolutions greatly improved the
storage, transmission, and retrieval of information, but the deepest
layer of interpretation, synthesis, and judgment still depended on human
cognition. Machines could execute explicitly defined procedures, but
they could not participate broadly in the flexible processing of
complex, open-ended information.

Large language models mark a qualitative change. Although they have not
yet reached human-level general intelligence, their breadth of
competence already makes them comparable to humans in many domains of
information processing. The significance of the current revolution,
therefore, is not simply that machines can calculate faster or search
more efficiently, but that machine information processing has crossed an
important threshold of complexity. This is why the AI revolution should
be understood as an unprecedented event in human history. If human
language accelerated the evolution of civilization relative to
biological evolution, then the era of Human-AI symbiosis may accelerate
the evolution of civilization on an even shorter timescale.

\subsection{The Fundamental Change Brought by the LLM
Revolution}\label{the-fundamental-change-brought-by-the-llm-revolution}

The most fundamental change brought by the LLM revolution is that human
know-how is becoming replicable and shareable at scale. This point is,
in our view, more important than any single application. Before AI, human
beings were already able to share explicit knowledge through books,
papers, formulas, software, and formal instructions. However, a large
part of real human capability does not exist in fully explicit form. It
exists as know-how: judgment built from experience, practical intuition,
habits of problem solving, tacit understanding of context, and the
ability to respond to subtle variations in real situations.

Traditionally, this kind of know-how could only be transmitted through
close human interaction. It required apprenticeship, repeated practice,
observation, correction, and often long periods of collaboration. A
textbook can explain principles, but it cannot fully convey how an
expert actually works. A paper can present a result, but it usually
omits the failed attempts, intermediate decisions, practical tricks, and
contextual understanding that were necessary to obtain that result. For
this reason, knowledge could be distributed broadly, while experience
remained difficult to copy. This gap has always limited the speed at
which human capability can spread.

Large language models begin to change this condition in a fundamental
way. By learning from large-scale records of human language and by
interacting directly with users in practical tasks, AI systems can
absorb and reproduce patterns of reasoning, explanation, decision
making, and problem decomposition that were previously embedded only in
human practice. They do not merely store explicit statements of
knowledge. They can also approximate the operational form of expertise:
how to approach a problem, how to ask the next question, how to organize
a workflow, and how to adapt general principles to a specific context.

This is why AI creates a new kind of productivity. Its deepest
contribution is not only automation in the narrow sense, but the
large-scale replication and distribution of human know-how. Capabilities
that once depended on direct personal teaching can now, at least in
part, be encoded into AI systems and shared widely. In this sense, AI
extends the social reach of expertise. It allows practical experience
that was once local, fragile, and difficult to transfer to become more
accessible, reusable, and combinable across individuals and
institutions.

This, we believe, is the core of the AI revolution. The central issue is
not only that machines can do tasks, but that they can carry and
transmit forms of human capability that were previously non-transferable
except through labor-intensive human teaching. Once know-how becomes
reproducible at scale, the organization of education, research,
production, and collaboration will all change. This is the most
fundamental mechanism by which AI generates new productive forces, and
it is the reason its impact reaches far beyond ordinary technological
progress.

\section{AI for Science}\label{ai-for-science}

Based on the observations summarized above, the AI revolution will bring
fundamental changes to all fields. Among its many applications, the most
important may be its impact on the way innovation itself takes place,
because this change has the potential to create the greatest long-term
value. For this reason, it is especially important to examine how AI
will affect science and technology, where the central task is to expand
the frontier of knowledge and generate genuinely new ideas. In earlier
stages of technological development, new tools improved the efficiency
of research, but they did not directly transform the process of
innovation itself. Large language models, by contrast, introduce the
possibility that AI may participate not only as a tool, but also as a
collaborator in scientific discovery. In this section, we will first
discuss the challenges in current scientific research, then examine the
opportunities brought by AI for Science, and finally consider the
challenges AI still faces and the next steps forward.

\subsection{Pain Points in Scientific
Research}\label{pain-points-in-scientific-research}

\begin{figure}
\centering
\includegraphics{fig2.jpeg}
\caption{Illustration of the major pain points in scientific research,
including the time cost of understanding prior work, the loss of tacit
knowledge, limits of collaboration, and administrative burden.}
\end{figure}

To understand what AI brings to research, we must review common problems
currently facing scientific enquiry. While challenges vary by field,
several are universal:

\begin{enumerate}
\def\labelenumi{\arabic{enumi}.}
\tightlist
\item
  \textbf{Time Costs:} Understanding industry progress and learning from
  others' work requires immense time.
\item
  \textbf{Loss of Tacit Knowledge:} A vast amount of ``intermediate''
  experience and data accumulated during research is not reflected in
  papers, forcing other scholars to explore from scratch.
\item
  \textbf{Collaboration Limits:} The scale of research collaboration is
  constrained by human communication costs, making large-scale and
  cross-disciplinary cooperation difficult.
\item
  \textbf{Administrative Burden:} Significant time is consumed by
  non-creative tasks, such as writing papers, peer review, writing grant
  proposals, and explaining work to others after the research is
  completed.
\end{enumerate}

As discussed in the background section, the deeper cause of these
problems is that explicit knowledge can be widely shared, while human
experience and know-how are much harder to transfer. In scientific
research, this distinction is especially important. A paper can record
the final result, the formal method, and selected evidence, but it
usually cannot fully capture the actual process by which the work was
completed. Many essential elements of research remain tacit: how to
choose a promising direction, how to avoid unproductive trials, how to
debug an experiment or a codebase, how to judge whether an unexpected
result is an error or a discovery, and how to adjust methods when
conditions depart from the ideal assumptions described in publications.
As a result, each new student or collaborator must spend substantial
time rebuilding practical understanding that already exists in the
community but has not been fully transmitted. This is one reason why
research training is slow, cross-disciplinary collaboration is
difficult, and many results remain hard to reproduce in practice. If
this layer of scientific know-how could be captured and shared more
effectively, the speed and structure of research collaboration would
change fundamentally. This is one of the most important new
possibilities opened by the AI revolution.

\subsection{Agentification of Scientific
Research}\label{agentification-of-scientific-research}

\begin{figure}
\centering
\includegraphics{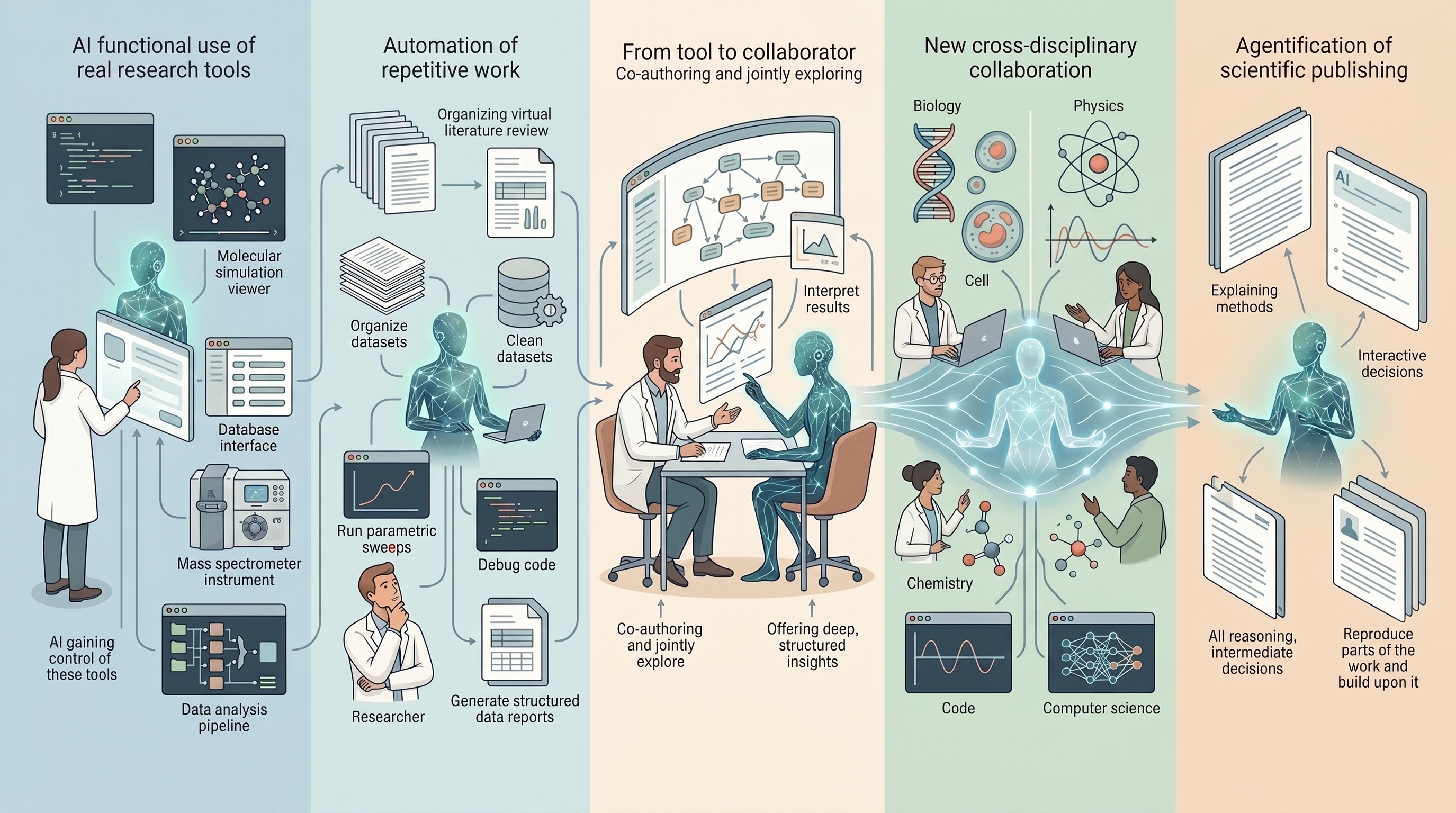}
\caption{Illustration of the agentification of scientific research, from
AI use of research tools and automation of repetitive work to scientific
collaboration, cross-disciplinary interaction, and agentic publishing.}
\end{figure}

The application of LLMs in science is already underway, with AI agents
assisting research in fields such as biology, mathematics, chemistry,
theoretical physics, and machine learning. Although current applications
are still at an exploratory stage, researchers in many areas have
already identified an increasing number of meaningful use cases
\citep{bubeck2025early, bran2024chemistrytools, wang2025clinical, li2024modeldiscovery, guevara2026graviton, brenner2026openproblem, lu2025theoreticalphysics}.
We believe that AI will ultimately bring a fundamental big change to
scientific research across disciplines. At the same time, this shift
will not arrive as a finished system designed in advance. It will be
discovered gradually by an open community in which human researchers and
AI systems learn to work together and reshape the research process step
by step.

We refer to this process, in which AI becomes progressively more deeply
involved in scientific work and thereby changes the structure of
research itself, as the {\bf agentification of scientific research}. The key
idea is not simply that AI becomes a better tool, but that it gains
increasing autonomy, continuity, and presence within the actual workflow
of science. In this subsection, we will discuss several dimensions of
this transformation. Roughly speaking, we will proceed from the near-term
and more concrete changes to the longer-term and more structural ones.

\paragraph{AI Use of Research Tools}\label{ai-use-of-research-tools}

The first step in AI participation in scientific research is to give AI
access to the tools that researchers actually use\citep{bran2024chemistrytools, schmidgall2025agentlab, shao2025sciscigpt, zhou2026paperprogram, deng2026physvec, liu2025vaspilot, zhu2025materialsgalaxy}. In theoretical work,
this includes computational software, simulation packages, coding
environments, databases, and computing resources. In experimental work,
it includes the software and control systems used to operate
instruments, collect data, adjust parameters, and monitor procedures.
Without such access, AI remains confined to the chatbox. It can provide
suggestions or explanations, but it cannot directly take part in the
real process of doing research.

Once AI can use research tools, its role changes qualitatively. It can
move from merely describing what should be done to actually carrying out
parts of the work. As model capabilities improve, this can begin with
simple and repetitive tasks, but it can gradually extend to longer and
more complex workflows, including the ability to respond to unexpected
situations. In this sense, tool use gives AI a ``physical body'' within
the world of scientific research. It allows AI to remain present
throughout the process rather than appearing only at isolated moments of
consultation.

This step is important not only because it improves efficiency, but also
because it is essential for the future development of AI itself. Once AI
is embedded in real workflows, it can encounter valuable forms of
scientific data and practical feedback that are absent from conventional
training corpora. These include failed trials, intermediate states,
instrument behavior, workflow decisions, and context-dependent
adjustments. Such data are crucial for building systems that can
participate meaningfully in frontier research rather than only in
textbook-level reasoning.

\paragraph{Automation of Repetitive
Work}\label{automation-of-repetitive-work}

After AI gains access to the necessary tools, the next priority is not
to ask immediately whether it can generate major new ideas. A more
practical and productive first step is to let it take over the routine
and repetitive parts of research. This is analogous to the role of a
beginning student. Human researchers also start by learning the workflow
of research through constrained and repeatable tasks before making more
independent contributions. AI should follow a similar path.

This stage includes literature review, organization of prior work,
implementation of standard pipelines, reproducible parts of theoretical
and experimental analysis, instrument debugging, targeted measurements,
parameter sweeps, data cleaning, and the preparation of routine reports
or summaries. These tasks are often time-consuming but not the main
source of originality. Automating them would immediately reduce the
burden on researchers and allow more human effort to be directed toward
judgment, interpretation, and creative thinking.

Equally importantly, this stage gives AI the opportunity to accumulate
experience across the full research process. Through participation in
routine work, AI can gradually learn how scientific projects actually
proceed, where bottlenecks arise, and how human researchers prefer to
collaborate. The most effective pattern of human-AI cooperation may
differ substantially across disciplines, and these patterns will likely
be discovered through practice rather than designed in advance. As AI
becomes more capable, the complexity of the tasks it handles can then
increase step by step.

\paragraph{From Tool to Collaborator}\label{from-tool-to-collaborator}

Building on the previous two stages, AI may eventually cross an
important threshold: the transition from tool to collaborator. This
threshold may be defined differently in different disciplines, but a
practical criterion is whether AI can make a contribution comparable to
that of a graduate student on a research project. If, in a concrete
piece of scientific work, the contribution of an AI system becomes
genuinely comparable to that of a human student co-author, then even if
the AI still has clear weaknesses relative to humans, it has already
entered the inner space of research rather than remaining an external
aid
\citep{gottweis2025coscientist, shao2025sciscigpt, schmidgall2025agentlab, brenner2026openproblem, deng2026physvec}.

This threshold is important because it marks a change in who
participates in defining the frontier of innovation. Before that point,
AI mainly helps humans execute or accelerate tasks that humans have
already formulated. After that point, AI begins to influence the
direction, structure, and content of scientific discovery itself. Cases
of AI providing useful suggestions, novel hypotheses, or unexpected
connections are already beginning to appear, and such cases may
gradually become more common.

Because AI capabilities are improving rapidly, reaching the collaborator
stage may also imply that surpassing human performance in some
dimensions is not far away. For this reason, one can think of
authorship-level scientific contribution as a new kind of Turing test
for AI for Science. The central question is not whether AI can imitate
human conversation, but whether it can participate in the production of
publishable scientific knowledge at a level comparable to a recognized
human contributor.

\paragraph{New Cross-Disciplinary
Collaboration}\label{new-cross-disciplinary-collaboration}

As AI becomes more capable, it may also lower the barriers between
disciplines. Many important scientific opportunities arise at the
boundary between fields, but such collaboration is often limited by
differences in language, method, background knowledge, and research
culture. AI can help translate concepts across fields, summarize
unfamiliar literature, connect tools and datasets, and reduce the
communication cost that normally prevents deeper collaboration.

This matters because cross-disciplinary work is often where major
breakthroughs occur, yet it is also where human coordination is most
difficult. A biologist, a physicist, and a machine learning researcher
may each hold part of the knowledge required for a new discovery, but
bringing these parts together takes substantial effort. AI can serve as
an active interface across such domains, making collaborations easier to
initiate and more productive to sustain\citep{zhang2025bohrium, chai2025scimaster}. 
AI's participation in such collaborations will also lead to the emergence of new collaboration
platforms and collaboration patterns, much as the World Wide Web and
arXiv.org transformed scientific communication and collaboration in
earlier eras.

\paragraph{Agentification of Scientific
Publishing}\label{agentification-of-scientific-publishing}

The agentification of scientific research may eventually extend to
scientific publishing itself. Today, scientific results are usually
presented through static papers, which compress a complex research
process into a limited and highly standardized form. As discussed
earlier, this format is efficient for archiving explicit knowledge, but
it is poorly suited to transmitting the full depth of scientific
know-how.

When AI is deeply involved in scientific research at the level of
collaborator, a natural consequence is a new format of publication:
instead of publishing only a paper, we may {\bf publish the agent itself}. A
scientific result could then be represented not only by a static
document, but also by an interactive AI agent capable of explaining the
background of the work, the methods used, the reasoning process,
intermediate decisions, and relevant tool interfaces. In addition, such
an agent could help reproduce parts of the work, rerun standard
analyses, or even extend the original project. For example, instead of
only reading a methods section, a reader could directly ask the research
agent why a certain approximation was chosen, what alternative routes
were considered, or how the conclusions would change under a modified
assumption. Current publications do not yet directly propose publishing
the research agent itself, but adjacent work on end-to-end AI research
systems, AI authorship and review, and publisher responses to AI in
scientific communication suggests that scientific outputs are already
moving toward more agent-mediated and interactive forms
\citep{lu2024aiscientist, bianchi2026agents4science, bertolo2024publishing, kusumegi2025production, liang2025usage}.

Such an agent could also adapt its explanation to different audiences.
It might provide a concise conceptual introduction for a student, a
technical explanation for a specialist, or an implementation-oriented
guide for a researcher trying to build on the result. In this sense,
publishing would become less like depositing a static record and more
like deploying a living interface to the research itself. This would
greatly reduce the loss of tacit knowledge that occurs when complex work
is compressed into a conventional paper format.
At the same time, publication must balance two different requirements:
the core scientific content must remain stable enough to be cited,
checked, and criticized, while the agent layer should be able to
explain that content in multiple interactive ways for different
audiences. The key question is therefore to find an appropriate
combination of a stable archival record and a flexible interpretive
interface.

Agentic publication could also accelerate the creation of new work based
on existing work. If a scientist wants to build on a previous result,
the relevant research agent could help identify assumptions, reproduce
calculations, locate useful datasets, or suggest possible extensions.
More interestingly, it opens the possibility of agent-agent
collaboration. If the authors of two different works believe that a new
discovery may lie at the intersection of their results, they could allow
the agents representing those works to exchange ideas, compare
assumptions, explore compatibility, and generate possible directions for
follow-up research. Human researchers could then evaluate and develop
the most promising possibilities. In this way, agentic publication would
not only improve scientific communication, but also create new
mechanisms for scientific exploration.

In academia, publication is not only a way to share results; it also
plays a central role in evaluation and reward. To a large extent,
academic achievement is still measured by the impact of papers. The
agentification of scientific publishing therefore implies that the
evaluation system itself may need to change fundamentally. This would
address a long-recognized problem in the research community. The current
academic publishing system has been in place for more than a century and
suffers from many widely known limitations. Peer review depends heavily
on the unpaid labor of scholars, while journals often charge substantial
fees to both authors and readers. At the same time, many contributions
that are highly valuable to the community, such as maintaining an
important open-source scientific software library, building a dataset,
or developing a shared experimental platform, are often undervalued
because they do not fit naturally into the standard paper format.

If publication becomes more agentic, it may create new ways to recognize
and evaluate such contributions. Although the final form of such a
system remains uncertain. What seems much clearer is that the evaluation
and reward structure of academia will change in a fundamental way. For
this reason, it is already valuable today to explore new platforms for
agentic publication that are lower-cost, more efficient, and more open
than the current system. As with many previous transformations in
science, the eventual form of such a platform will likely emerge from
the collective experimentation of the research community.

\subsection{Challenges in AI for
Science}\label{challenges-in-ai-for-science}

\begin{figure}
\centering
\includegraphics{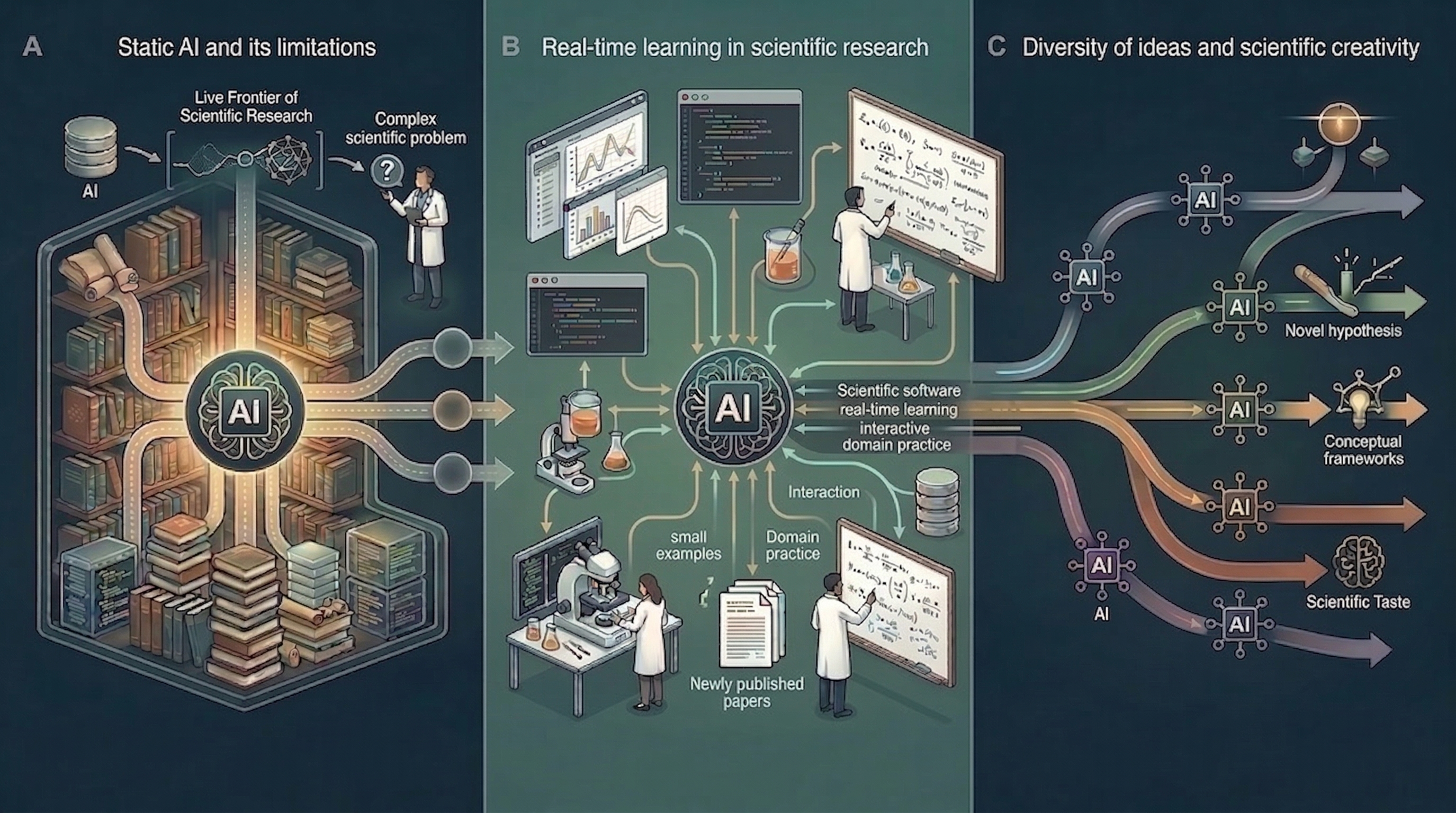}
\caption{Illustration of why the next step for AI in science is
real-time learning and diversity of ideas, enabling continuous
adaptation to frontier research and more original scientific discovery.}
\end{figure}

The opportunities described above are substantial, but they should not
be confused with fully realized capabilities. To move from promising
demonstrations to a genuine transformation of scientific research,
several core challenges still need to be addressed:

\begin{enumerate}
\def\labelenumi{\arabic{enumi}.}
\tightlist
\item
  \textbf{Lack of Frontline Data:} While models excel at textbook-level
  problems, they struggle with real research scenarios because training
  data does not cover the minute details of every vertical niche.
  Experts must lead AI in research to expose it to real data for
  specialized training.
\item
  \textbf{Lack of Real-Time Updates:} In scientific research, new tools
  and concepts are constantly being invented, which cannot be rapidly
  mastered by models through training. AI needs the ability to learn
  continuously. Currently, context engineering protocols like the
  Model Context Protocol (MCP)\cite{modelcontextprotocol} and Agent Skills Protocol\cite{agentskillsprotocol} are addressing this need by connecting AI to tools and knowledge.
\item
  \textbf{Need for New Evaluation Frameworks:} Current evaluation
  methods are still largely based on benchmark-style testing. This
  approach has two important limitations. First, existing benchmarks
  tend to focus on broad and relatively mainstream domains, so they are
  often not precise enough for highly specialized research areas.
  Second, benchmarks are usually question-answer style evaluations,
  which are poorly suited to measuring how an agent performs in
  long-term scientific collaboration. Recent domain-specific evaluation
  efforts already point toward the kind of richer assessment that will
  be needed, including long-context scientific reasoning tasks,
  expert-built condensed-matter theory problems, expert-graded
  literature understanding, and end-to-end verified physics workflows
  \citep{cui2025curie, pan2025cmtbenchmark, guo2025expertworldmodels, deng2026physvec,wang2025qmbench,wang2025cmphysbench,deng2026towards}.
  Once AI begins to function as a research collaborator, it may need to
  be evaluated more like a graduate student: not only by whether it can
  produce correct answers in isolated tasks, but by how it performs over
  time in real projects, how well it learns from feedback, how reliably
  it contributes to a workflow, and whether its judgments are useful in
  context. How to provide this kind of long-horizon, high-resolution
  feedback, and how to incorporate it effectively into AI training,
  remains an open problem.
\end{enumerate}

Among these challenges, we believe the most important next step for AI
development is the ability to learn in real time\citep{wu2024continual}. Compared with humans,
current AI systems still rely on too much data and too much time to
acquire new capabilities through training. Human researchers, by
contrast, can often learn a great deal from a small number of examples,
a brief discussion, or a limited amount of direct experience. If AI is
to become a deeper participant in scientific work, it must move closer
to this mode of online learning: absorbing new tools, concepts,
feedback, and domain-specific practices continuously during real tasks
rather than only through slow training cycles.

This capability matters not only because it would make AI more powerful,
but also because it is essential for {\bf diversity of ideas}. Creative
scientific work depends on more than competence alone. It also depends
on differences in perspective, interest, and taste across researchers.
In physics, for example, there are tens of thousands of researchers with
distinct intuitions about what problems are important, what methods are
promising, and what anomalies are worth pursuing. When a new opportunity
for discovery appears, the person who makes the crucial leap is often
distinguished not only by capability, but also by intellectual taste and
interest. For this reason, a research community needs sufficient
diversity of ideas in order to sustain original discovery.

Current AI models clearly lack this kind of diversity
\citep{wang2025divergent, hao2026focus}. Although
they can be prompted to display different personalities, their views on
the same problem are often highly similar. Without a mechanism for
continuous and diverse learning, AI systems will tend to reproduce the
dominant patterns already present in their training data. That
limitation would make truly creative work difficult. For this reason,
online learning is not simply a technical improvement; it is a central
requirement for AI to become a genuine collaborator in scientific
discovery. An interesting open question is whether achieving meaningful
diversity in AI requires a new architecture, or whether a sufficiently
strong improvement in-context learning would already be enough.

\section{Conclusion}\label{conclusion}

The central argument of this article is that the fundamental
significance of the AI revolution is not simply automation, nor the
acceleration of information retrieval, but a deeper change in the
dynamics of information itself. In earlier stages of history, major
transformations occurred when new carriers of complex information
emerged: first life, with DNA and RNA, and later human civilization,
with language. The current AI revolution marks another such
transformation. For the first time, complex information processing is no
longer limited to the human brain. More importantly, large language
models begin to make human know-how, not only explicit knowledge,
increasingly replicable and shareable. This is the deepest source of the
new productivity created by AI, and it is the reason this revolution may
eventually reshape the structure of human civilization.

From this perspective, AI for Science is important not only because
science is one domain among many, but because the transformation of
scientific innovation has exceptional significance for both AI and
science itself. For science, the key promise of AI is that it may reduce
the cost of transmitting know-how, accelerate collaboration, and
eventually change the very process by which new ideas are generated,
tested, and communicated. For AI, science is one of the most demanding
environments in which real progress can be made. Scientific research
exposes AI to frontier problems, specialized tools, real feedback, and
open-ended collaboration. In this sense, AI for Science is not merely an
application of AI; it is also a path for developing more capable AI.

The roadmap of AI for Science, as argued here, is gradual. It begins
with giving AI access to the tools of research, so that it can move from
the chat interface into the real workflow of science. It then proceeds
through the automation of routine and repetitive work, where AI can
accumulate practical experience while reducing the burden on human
researchers. On this basis, AI may eventually cross the threshold from
tool to collaborator, making contributions comparable to those of a
human student or co-author. Beyond that, AI may help create new forms of
cross-disciplinary collaboration and ultimately reshape the way
scientific work is published and evaluated. This process of
progressively deeper participation is what we have named as the
agentification of scientific research.

One of the most important consequences of this process is the
possibility of agentic publishing. If AI becomes a genuine collaborator
in research, then it is natural that scientific publication may evolve
from static papers toward interactive research agents. Such agents could
preserve more of the reasoning, intermediate decisions, methods, and
practical know-how that are currently lost in conventional publication.
They could also make scientific results more accessible to different
audiences, accelerate follow-up work, and even enable new forms of
agent-agent exploration across related works. For this reason, agentic
publishing is not a secondary issue. It is closely connected to the
future of scientific communication, academic evaluation, and the reward
structure of research itself.

At the same time, the article has emphasized that the future of AI for
Science does not depend on capability alone. A fundamental requirement
for genuine scientific creativity is diversity of ideas. In human
research communities, discovery depends not only on knowledge and skill,
but also on differences in perspective, interest, taste, and judgment.
These differences are essential because they allow the community to
explore many directions at once and to recognize opportunities that
would otherwise be ignored. Current AI systems still lack this kind of
diversity. Without the ability to learn continuously from varied
contexts and communities, they will tend to reproduce the dominant
patterns already present in their training data. For this reason, online
learning and mechanisms for preserving diversity of ideas are not
peripheral technical issues. They are central to whether AI can become a
true collaborator in original scientific discovery.

The overall conclusion, therefore, is that AI for Science should be
understood as both a scientific and a civilizational project. Its goal
is not merely to make existing research faster, but to build a new
paradigm in which human researchers and AI agents jointly participate in
the production, transmission, and evaluation of knowledge. The path
toward this future will require new tools, new evaluation frameworks,
new collaboration platforms, and new publishing systems. It will also
require an open community in which human experience is continuously
taught to AI through real work. If this path succeeds, the most profound
impact of AI may be that it changes not only what we know, but how
humanity creates new knowledge at all.

\section{Acknowledgement}\label{acknowledgement}

This article is a summary of the author's thoughts about the AI
revolution since 2023. Some earlier discussion of these ideas can be
found in an earlier article \citep{qi2024timeinfo} and some lectures such as
\citep{qi2025pirsa, qi2025youtube}. An earlier version of this article
was also posted on the \url{https://ai4.science} discussion forum
\citep{qi2026forum}. The author would like to acknowledge helpful
discussions with Diane Greene, Chaoxing Liu, Sirui Lu, Chen Nie, Xiaodong Xu, Binghai
Yan, Bojun Yan, and Yi-Zhuang You on related topics. The author used large language models, primarily GPT-5.4 via Codex, to assist in polishing the manuscript. The figures were generated using Gemini app.

\bibliography{references.bib}

\end{document}